\documentclass[fleqn,10pt]{wlscirep}
\usepackage[utf8]{inputenc}
\usepackage[T1]{fontenc}
\usepackage{lineno}
\usepackage{float}
\usepackage{multirow}
\usepackage{xcolor}
\usepackage{subfigure}
\usepackage{CJKutf8}
\usepackage{soul}
\usepackage{color}
\newcommand{\zh}[1]{\begin{CJK*}{UTF8}{gbsn}#1\end{CJK*}}
\newcommand{\dataset}{HUST-OBC}

\definecolor{green}{RGB}{112, 173, 71} 

\title{An open dataset for oracle bone character recognition and decipherment}

\author[1,$\dag$]{Pengjie Wang}
\author[1,$\dag$]{Kaile Zhang}
\author[2]{Xinyu Wang}
\author[3]{Shengwei Han}
\author[3]{Yongge Liu}
\author[1]{Jinpeng Wan}
\author[1]{Haisu Guan}
\author[1]{Zhebin Kuang}
\author[4]{Lianwen Jin}
\author[1]{Xiang Bai}
\author[1,*]{Yuliang Liu}
\affil[1]{Huazhong University of Science and Technology, Wuhan, 430074, China}
\affil[2]{The University of Adelaide, SA, Adelaide, 5005, Australia}
\affil[3]{Anyang Normal University, Anyang, 455000, China}
\affil[4]{South China University of Technology, Guangzhou, 510641, China}

\affil[*]{Corresponding author(s): Yuliang Liu (ylliu@hust.edu.cn)}

\affil[$\dag$]{These authors contributed equally to this work}

\begin{abstract}
    Oracle bone script, one of the earliest known forms of ancient Chinese writing, presents invaluable research materials for scholars studying the humanities and geography of the Shang Dynasty, dating back 3,000 years. The immense historical and cultural significance of these writings cannot be overstated. However, the passage of time has obscured much of their meaning, presenting a significant challenge in deciphering these ancient texts. With the advent of Artificial Intelligence (AI), employing AI to assist in deciphering Oracle Bone Characters (OBCs) has become a feasible option. Yet, progress in this area has been hindered by a lack of high-quality datasets. To address this issue, this paper details the creation of the \dataset{} dataset. This dataset encompasses 77,064 images of 1,588 individual deciphered characters and 62,989 images of 9,411 undeciphered characters, with a total of 140,053 images, compiled from diverse sources. The hope is that this dataset could inspire and assist future research in deciphering those unknown OBCs. All the codes and datasets are available at \href{https://github.com/Pengjie-W/HUST-OBC}{https://github.com/Pengjie-W/HUST-OBC}.
\end{abstract}

\begin{document}

\flushbottom
\maketitle
\thispagestyle{empty}

\section*{Background \& Summary}

\begin{figure}[h!]
    \centering
    \includegraphics[width=0.8\linewidth]{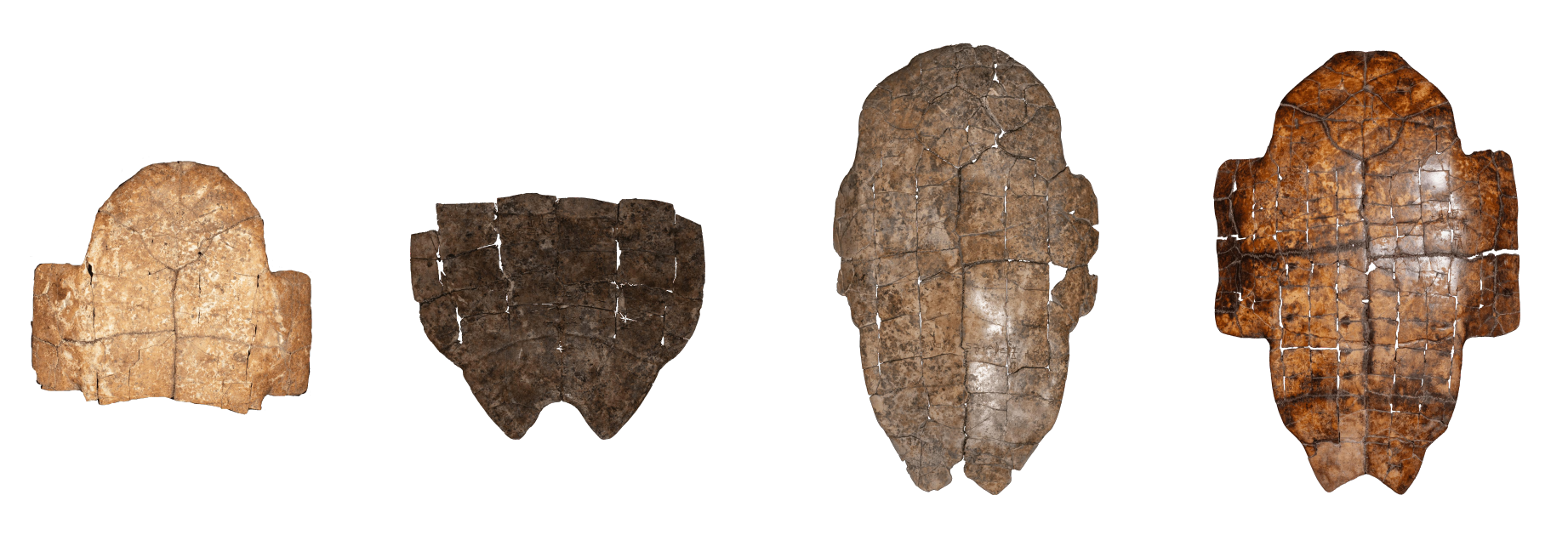}
    \caption{3000-year-old Shang Dynasty oracle bones. They were all unearthed at the Huayuanzhuang East in Yinxu, China, and are currently housed at the Anyang workstation of the Institute of Archaeology, Chinese Academy of Social Sciences. These oracle bones date back to the reign of King Wu Ding of the Shang Dynasty.}
    \label{fig:oracle-bone}
\end{figure}
Oracle bone script, etched onto turtle shells and animal bones, stands as one of the earliest forms of writing discovered in China (see Figure~\ref{fig:oracle-bone}). These inscriptions, dating back about 3,000 years, offer a window into the human geography of the Shang Dynasty (1600 BCE - 1046 BCE), an ancient feudal dynasty ruled in the Yellow River valley. The content encompasses a range of topics including astrology, meteorology, animal husbandry, religion, and ritual practices~\cite{boltz1986early, keightley_1979}. Similar to other ancient scripts, the meanings of many Oracle Bone Characters (OBCs) have been lost over time. Of the 160,000 pieces unearthed, they reveal more than 4,600 distinct OBCs, yet only about a thousand of these have been deciphered, with their meanings and corresponding modern Chinese characters confirmed~\cite{bazerman2009handbook}.

The target of deciphering these ancient inscriptions is to translate OBCs into modern Chinese, with characters corresponding one-to-one with the same meaning. However, the deciphering task at the character level is complicated by several factors. Historically, the methods of preservation and excavation were not always ideal, leading to many of the oracle bones being damaged. This damage often results in partial, unclear, or illegible inscriptions, making interpreting them arduous. Therefore, most of the images used in current Oracle Bone Character (OBC) research are scanned images that have undergone denoising and other processing or artificially transcribed images. In addition, the nature of oracle bone script as an early writing system means that it underwent significant evolution. There is a considerable variation in the form of characters, with many characters appearing in multiple, sometimes radically different forms~\cite{gao2020distinguishing} but corresponding to the same Chinese character. This variability adds another layer of complexity to the deciphering process. All these factors contribute to making the full understanding of OBC not only challenging but also a rare feat, attracting the keen interest of scholars and historians alike in the field of ancient Chinese studies.

The past decades have witnessed the widespread applications of Artificial Intelligence (AI) in various fields. Notably, the immense success of handwritten text recognition (HTR)\cite{lecun1989handwritten,graves2008offline,bhunia2019handwriting} technology in processing modern texts has sparked interest in the potential use of AI to aid in deciphering OBCs. Modern AI algorithms, particularly those centered around deep learning models like artificial neural networks, typically require an extensive volume of data for training. This approach enables them like AlphaGo who defeated the world champion in a Go game\cite{silver2016mastering} to achieve, and sometimes surpass, human-level performance in specific tasks. A fundamental step towards employing these models to decipher OBCs involves the creation and annotation of a comprehensive high-quality dataset of OBCs. In the dataset, each OBC is labeled by its modern counterpart, while different OBCs with the same label are referred to as a category in the dataset. There have been some pioneering efforts in this area. For instance, Li et al.~\cite{HWOBC} built the HWOBC dataset by engaging experts from diverse academic backgrounds to handwrite OBCs. Fu et al.~\cite{OBI100} and Yue et al.~\cite{OBI125} subsequently proposed the OBI-100 and OBI-125 datasets, with the OBC images collected from books related to OBC research. Additionally, Guo et al.~\cite{20K} collected more than 20k OBC images from various websites to build the Oracle-20k dataset. These efforts lay a solid foundation for digitalization and recognition research of OBC. In addition, Li et al.\cite{li2024oracleboneinscriptionsmultimodal} created the Oracle Bone Inscriptions Multi-modal Dataset (OBIMD), but this dataset focuses on entire rubbings and lacks rich data on individual OBCs. However, the fact that these datasets have certain limitations also hinders their use in AI-assisted OBC decipherment:

\begin{itemize}
    \item They often have limited categories and samples of OBCs due to data collection from a single source.
    \item The annotation of the categories might not be deduplicated. As shown in Table~\ref{tab:Merging}, the same OBCs are categorized into different classes. 
    \item The lack of cross-validation from multiple sources casts doubt on the accuracy of some data.
    \item The datasets comprise only deciphered oracle bone images, making them unsuitable for deciphering tasks.
    \item Some datasets contain unprocessed images, filled with noise or blur.
\end{itemize}

To address these issues, we propose the high-quality \dataset{}\cite{HUSTOBC} dataset. The \dataset{} dataset was collected from three different sources, including books, websites, and existing databases. \dataset{} includes two types of OBC sample images: a) OBC images obtained from processed scans of rubbings of the original oracle bones; and b) handwritten OBC images based on the original oracle bones, further subdivided into traced images based on rubbings and manually drawn images based on the glyphs. To build \dataset{}, we designed a semi-automatic pipeline that collects and annotates data from various sources and had OBC experts review the dataset. As shown in Table~\ref{tab:Compare}, \dataset{} contains over 10k deciphered and undeciphered OBC categories and more than 140k images, making it one of the largest datasets for OBCs recognition and deciphering to date. We hope \dataset{} will aid and inspire future AI-assisted OBC research.

\begin{table}[t!]
\centering
\begin{tabular}{c|c|cc|cc}
\hline
\multirow{2}{*}{Dataset} & \multirow{2}{*}{Image Source} & \multicolumn{2}{c|}{Deciphered} & \multicolumn{2}{c}{Undeciphered} \\ \cline{3-6} 
 & & \#Characters & \#Images & \#Characters & \#Images \\ \hline
HWOBC & Book & 1,566 & 33,555 & 2,315 & 49,690 \\
OBI-100 & Book & 100 & 4,748 & 0 & 0 \\
OBI-125 & Book & 125 & 4,257 & 0 & 0 \\
Oracle-20k & Web & 261 & 20,039 & 0 & 0 \\ \hline
HUST-OBC & Book, Web, Database & 1,588 & 77,064 & 9,411 & 62,989 \\ \hline
\end{tabular}
\caption{Comparison of the scale of \dataset{} with other oracle bone character datasets.}
\label{tab:Compare}
\end{table}

\section*{Methods}

To construct a diverse dataset, we gathered images of OBCs from three distinct sources: \textit{book}, \textit{website}, and \textit{database}. To organize and merge data from these varied origins, as shown in Figure~\ref{fig:flowchart}, we designed a semi-automated pipeline comprising four key steps: \textit{Data Acquisition}, \textit{Automatic Annotation}, \textit{Data Integration}, and \textit{Data Validation}. In this section, we will delve into the details of each step.

\subsection*{Data Acquisition}

OBCs were inscribed on turtle shells and animal bones and buried underground for over 3,000 years. These precious artifacts are dispersed in museums and private collections worldwide, where they are meticulously preserved, making direct access to the text inscribed on the original oracle bones quite challenging. Thankfully, most of the publicly available oracle bones have been transcribed by experts, making them accessible in various forms for scholarly research. Specifically, for most authoritative books or websites, the images are processed or traced based on rubbings of oracle bones by experts. Building on this, HWOBC hired experts to manually draw each oracle bone character glyph, thus expanding the dataset with handwritten OBCs. As illustrated in Table~\ref{tab:data-source}, the \dataset{} dataset is constructed by gathering data from these diverse sources.

\begin{figure}[t!]
    \centering
    \includegraphics[width=\textwidth]{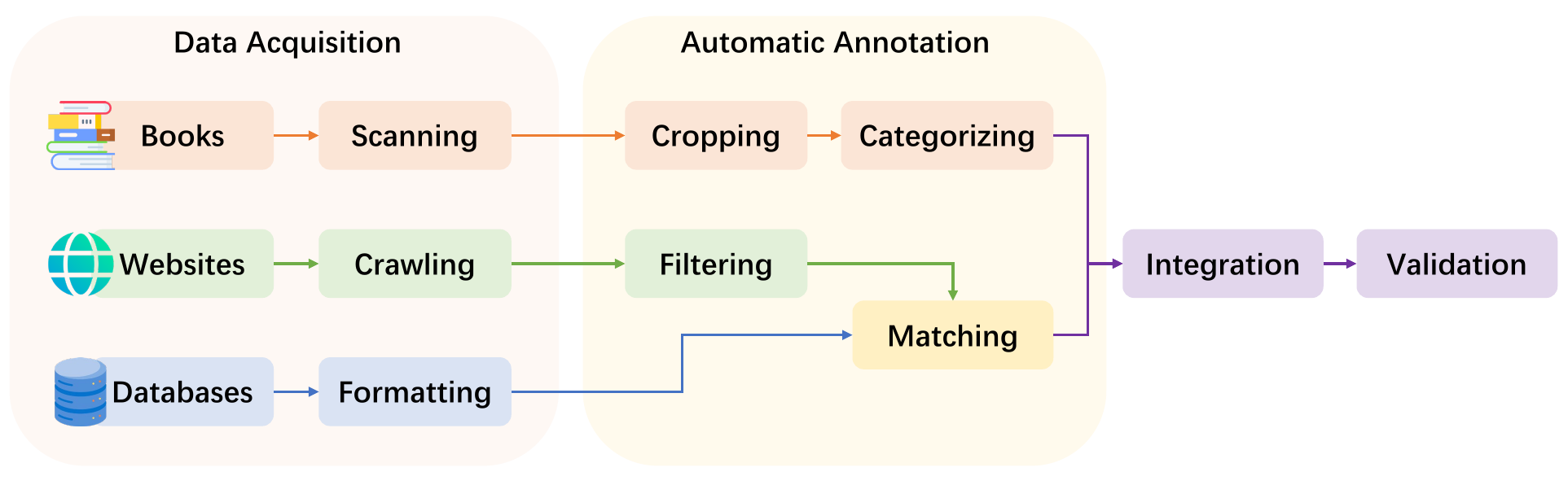}
    \caption{Flowchart of building the \dataset{} dataset.}
    \label{fig:flowchart}
\end{figure}

To ensure the diversity of the dataset, the HUST-OBC was built using data collected from various sources, including books, websites, and databases. As shown in Figure~\ref{fig:flowchart}, we designed specific pipelines for each data source to process and extract OBC images and their corresponding labels, detailed as follows.

\subsubsection*{Books}

Books remain the predominant form in documenting OBCs, with most discovered characters to date collected and interpreted in volumes like the \textit{New Compilation of Oracle Bone Scripts}, which ensures accuracy by incorporating the latest research in this field. Specifically, we utilized the following books as data sources while constructing the \dataset{} dataset.

\textbf{A. New Compilation of Oracle Bone Scripts} (\zh{新甲骨文编} \href{https://books.google.com/books?id=S0RergEACAAJ}{https://books.google.com/books?id=S0RergEACAAJ})\cite{X} encompasses samples of OBCs found since its initial discovery, as presented in all public materials.

\textbf{B. Oracle Bone Script: Six Digit Numerical Code} (\zh{甲骨文六位数字码检索字库} \href{https://books.google.com/books?id=pgvaxQEACAAJ}
{https://books.google.com/books?zid=pg vaxQEACAAJ})\cite{L} assigns digital codes to OBCs, annotating each code with its corresponding oracle bone character, modern Chinese character form, provenance, and other relevant details.

Since these books do not provide electronic databases or original image data, we manually scanned the pages of these books, obtaining 1,054 and 700 pages from books A and B, respectively. An example of scanned pages is presented in Figure~\ref{fig:book}(a).

\subsubsection*{Websites}

\begin{figure}[t!]
    \centering
    \includegraphics[width=\textwidth]{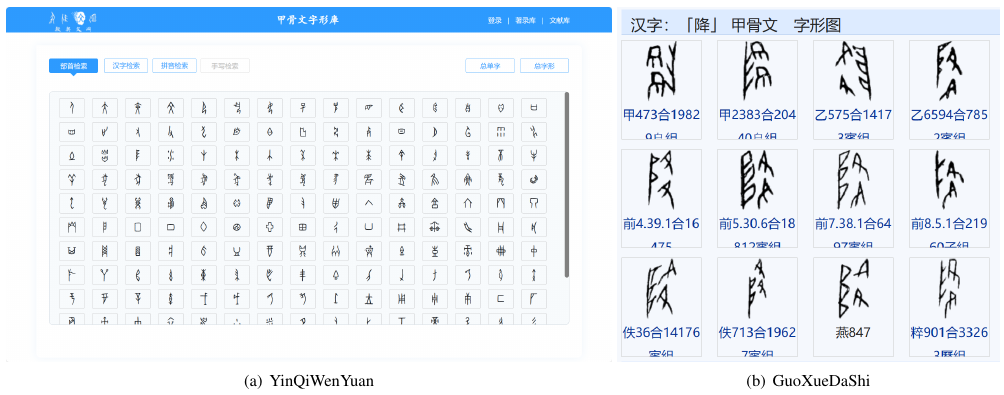}
    \caption{Screenshots of example website pages from YinQiWenYuan and GuoXueDaShi.}
    \label{fig:websites}
\end{figure}

With the widespread adoption of the Internet, websites have emerged as an alternative for hosting oracle bone data, offering more convenient retrieval capabilities. We have designed a web crawler program to collect data from the following websites:

\textbf{C. GuoXueDaShi} (\zh{国学大师} \href{https://www.guoxuedashi.net/jgwhj/}{https://www.guoxuedashi.net/jgwhj/}), initiated and maintained by enthusiasts of Chinese classical studies, which includes various historical texts including dictionaries, histories, etc. A screenshot of the website is shown in Figure~\ref{fig:websites}(b).

\textbf{D. YinQiWenYuan} (\zh{殷契文渊} \href{https://jgw.aynu.edu.cn}{https://jgw.aynu.edu.cn}) is a data platform maintained by the Key Laboratory of Oracle Bone Inscriptions Information Processing, archives various types of data, including photos of the original oracle bones, transcribed characters, and related research articles. A screenshot of the website is shown in Figure~\ref{fig:websites}(a).

These websites feature well-organized collections of OBC images, which have been meticulously scanned, cropped, and aligned. They are systematically categorized across various web pages, facilitating the use of web crawler technology to download these images in a categorized format efficiently.

\subsubsection*{Databases}

In recent years, the digitalization of ancient manuscripts and advancements in handwritten text recognition technology have opened new avenues in the study of OBC, which has led to the proposal of relevant datasets. We have included the following databases in \dataset{}.

\textbf{E. HWOBC} (\href{https://jgw.aynu.edu.cn/home/down/detail/index.html?sysid=2}{https://jgw.aynu.edu.cn/home/down/detail/index.html?sysid=2}) is a database specifically designed for the study and recognition of handwritten OBCs~\cite{HWOBC}. Compared to other books and websites that process or trace rubbings, to obtain more extensive handwritten samples of OBCs, the HWOBC dataset hired experts to manually draw each character glyph on a 400x400 pixel white background using a PC or smartphone, and then upload them to create a richer set of 83,245 handwritten OBC images.
% \begin{figure}[t!]
%     \centering
%     \subfigure[Scanned Book Page]{\includegraphics[height=6cm]{figures/scanned-page.jpg}\label{fig:scanned-page}}
%     \subfigure[Page Cropping]{\includegraphics[height=5.9cm]{figures/cropped-page.jpg}\label{fig:cropped-page}}
%     \caption{Extraction of OBC images from books. \textit{New Compilation of Oracle Bone Scripts} (left)}\cite{X} and \textit{Oracle Bone Script: Six Digit Numerical Code} (right)\cite{L}
%     \label{fig:book}
% \end{figure}
\begin{figure}[t!]
    \centering
    \includegraphics[width=\textwidth]{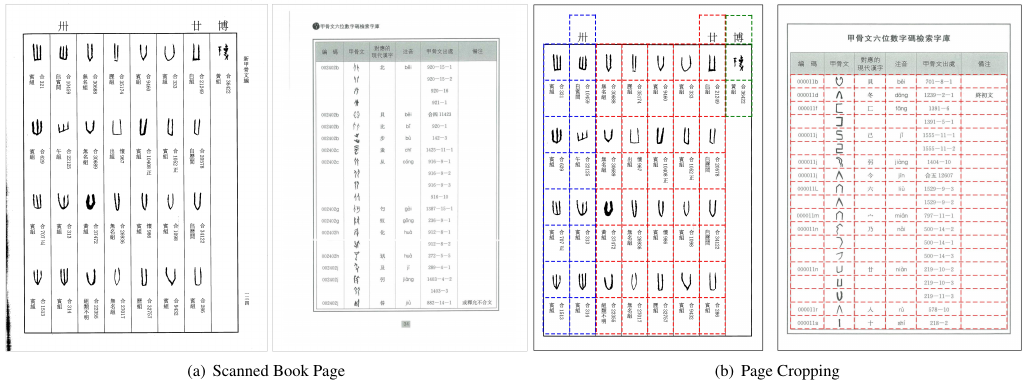}
     \caption{Extraction of OBC images from books. \textit{New Compilation of Oracle Bone Scripts} (left)\cite{X} and \textit{Oracle Bone Script: Six Digit Numerical Code} (right)\cite{L}.}
    \label{fig:book}
\end{figure}
% \begin{figure}[t!]
%     \centering
%     \subfigure[YinQiWenYuan]{\includegraphics[height=6cm]{figures/yqwy.png}\label{fig:yqwy}}
%     \subfigure[GuoXueDaShi]{\includegraphics[height=6cm]{figures/gxds.png}\label{fig:gxds}}
%     \caption{Screenshots of example website pages from YinQiWenYuan and GuoXueDaShi.}
%     \label{fig:websites}
% \end{figure}
\subsection*{Automatic Annotation}

Through data acquisition, we have gathered raw data from diverse sources. However, this data, in its current format, is not immediately usable. Hence, it necessitates further processing, including tasks such as cropping, annotating, and filtering.

\subsubsection*{Books}

The raw data for the books consists of scanned images of pages, each displaying several OBCs along with their corresponding annotations in modern standard Chinese. As shown in Figure~\ref{fig:book}(a), despite differences in the layout of the \textit{New Compilation of Oracle Bone Scripts} (left) and \textit{Oracle Bone Script: Six Digit Numerical Code} (right), they both employ a table-like vertical format. This arrangement facilitates the use of computer vision algorithms, such as edge detection, to automatically extract content from these pages. Specifically, as shown in Figure~\ref{fig:book}(b), we employed edge detection and other techniques from the OpenCV toolkit~\cite{bradski2000opencv} to crop the original scanned images by oracle bone characters, thereby obtaining individual slices of these characters. These slices are then categorized according to the layout rules, with each assigned a corresponding category ID. For example, as illustrated on the left side of Figure~\ref{fig:book}(b), the top of each column in the book is marked with the modern Chinese character equivalent to the OBC. If a column lacks such a marking, it implies that it belongs to the same category as the adjacent column on its right. In the figure, we used different colors of dashed lines to distinguish between categories. Using this method, we extracted 24,558 and 14,053 OBC images from source A and B books, respectively.

Although the slices are grouped during the cropping process, the corresponding modern Chinese character of each category remains unknown. A straightforward solution to determine the specific characters for each category is to use OCR techniques to recognize the marks in the books. However, most of the off-the-shelf OCR engines were trained only on commonly used modern Chinese characters and struggled to recognize the uncoded \textit{Liding} and unknown characters that may be presented in these books. To address this issue, we trained a category assigner (see Figure~\ref{fig:ocr}) to automatically identify these labels. The specific training procedures are detailed as follows:

\begin{figure}[t!]
    \centering
    \includegraphics[width=0.8\linewidth]{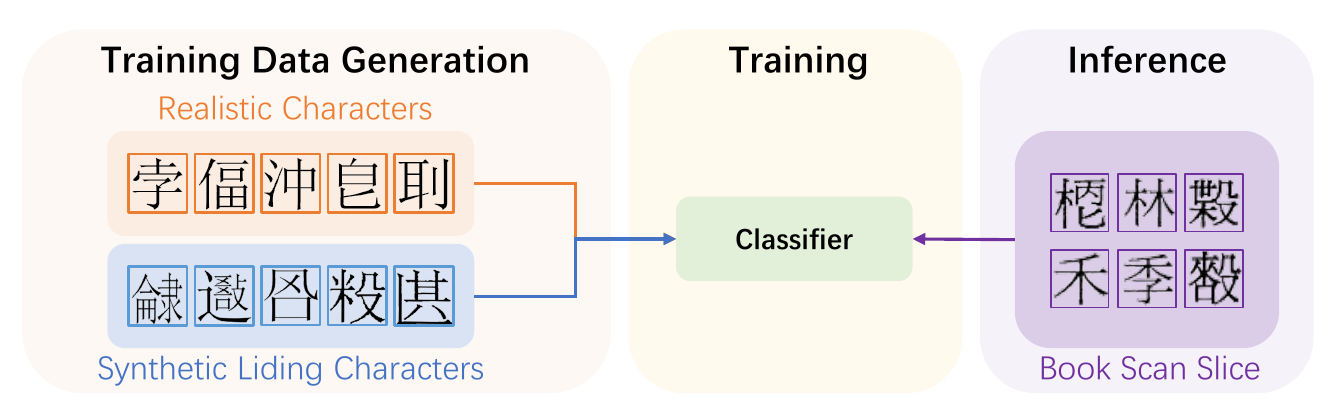}
    \caption{Schematic of the proposed category assigner.}
    \label{fig:ocr}
\end{figure}

\begin{enumerate}

    \item\textbf{Training Data Generation:} The Chinese character labels we need to identify are all in standard print typeface (as seen in the Chinese characters on the outside of the table at the top of Figure~\ref{fig:book}(b), left side), and each cropped image contains only a single, individual character. Thus, we can conveniently generate corresponding training samples using a similar SimSun font. As shown in the block on the left side of Figure~\ref{fig:ocr}, we generated font images for all realistic Chinese characters according to the Ideographic Description Sequence (IDS) and assigned each a unique category ID. Additionally, to address the recognition of uncommon characters that may appear in books, we randomly synthesized \textit{Liding} text using components like radicals of Chinese characters to serve as the \textit{Liding} category for training purposes. Practically, we generated one image for each of the total 88,899 Chinese characters included in the IDS and randomly synthesized $\alpha$ Liding character images in each training epoch.

    \item\textbf{Training:} Since each sample image contains only a single character, it is sufficient to train a simple classifier for recognition. For this purpose, we employed ResNet-50~\cite{resnet} as the backbone network to train the classifier. Additionally, we utilized a weighted balanced cross-entropy loss $L$ to address the issue of the imbalance in the number of training samples across different categories:
        \begin{equation}
            L=-\frac{1}{N+1}\left[\sum_{i=1}^N y_i \log \left(p_i\right)+\frac{1}{\alpha} y_{N+1} \log \left(p_{N+1}\right)\right]
        \end{equation}
    where $N$ is the number of categories, $P_{i}$ and $Y_{i}$ respectively represent the probabilities of the predicted and true labels being the $\text{i}^{th}$ category, taking values of 1 or 0, and $\alpha$ is the number of synthesized \textit{Liding} samples in each training epoch.
    \item\textbf{Inference:} During the inference phase, we input the Chinese character label images, which are cropped from the books, into the classification model trained in the second step. This process helps us determine the corresponding Chinese character for each category ID or whether it is an uncoded \textit{Liding} character.
\end{enumerate}

After completing the aforementioned procedures, all OBC images contained within the scanned pages of sourced books acquired during the data acquisition phase have been automatically extracted and accurately categorized according to their respective classes.

\begin{table}[t!]
    \centering
    \begin{tabular}{cccc}\hline
        Modern Character Category & Oracle Bone Character Images & Variant Characters Category & Oracle Bone Character Images \\\hline
        \includegraphics[width=1cm]{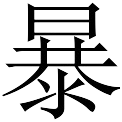}
        &\includegraphics[width=1cm]{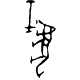}
        \includegraphics[width=1cm]{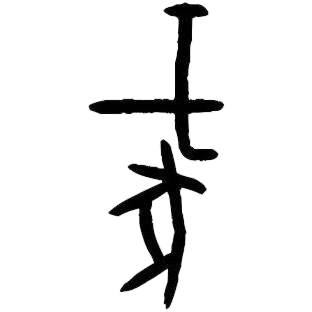}
        \includegraphics[width=1cm]{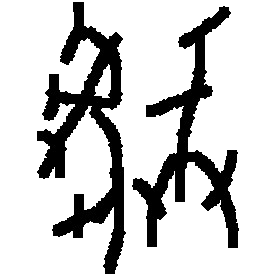}
        &\includegraphics[width=1cm]{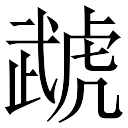}
        &\includegraphics[width=1cm]{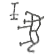}
        \includegraphics[width=1cm]{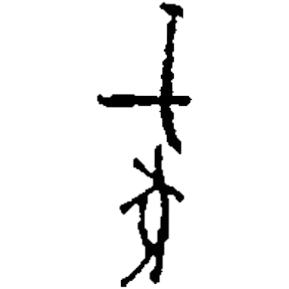}
        \includegraphics[width=1cm]{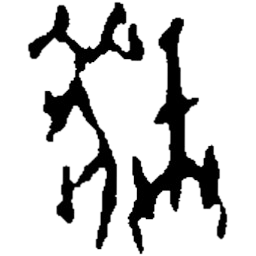}\\\hline
        \includegraphics[width=1cm]{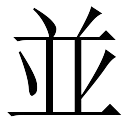}
        &\includegraphics[width=1cm]{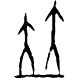}
        \includegraphics[width=1cm]{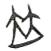}
        \includegraphics[width=1cm]{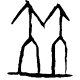}
        &\includegraphics[width=1cm]{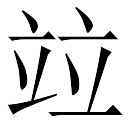}
        &\includegraphics[width=1cm]{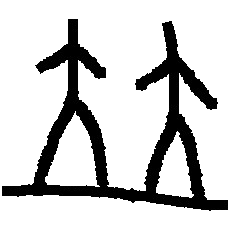}
        \includegraphics[width=1cm]{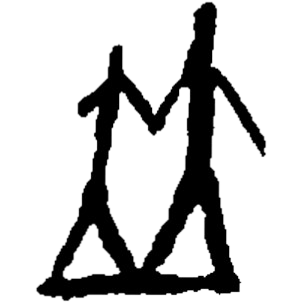}
        \includegraphics[width=1cm]{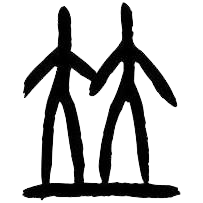}\\\hline
        \includegraphics[width=1cm]{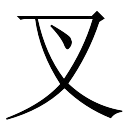}
        &\includegraphics[width=1cm]{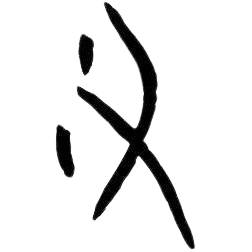}
        \includegraphics[width=1cm]{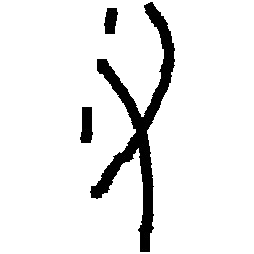}
        \includegraphics[width=1cm]{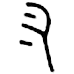}
        &\includegraphics[width=1cm]{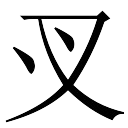}
        &\includegraphics[width=1cm]{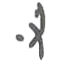}
        \includegraphics[width=1cm]{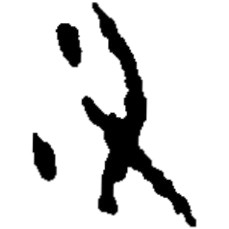}
        \includegraphics[width=1cm]{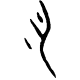}\\\hline
        \includegraphics[width=1cm]{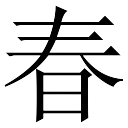}
        &\includegraphics[width=1cm]{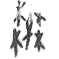}
        \includegraphics[width=1cm]{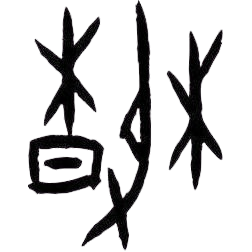}
        \includegraphics[width=1cm]{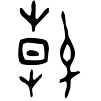}
        &\includegraphics[width=1cm]{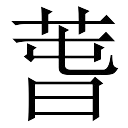}
        &\includegraphics[width=1cm]{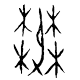}
        \includegraphics[width=1cm]{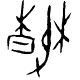}
        \includegraphics[width=1cm]{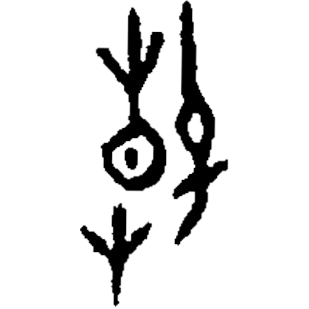}\\\hline
        \includegraphics[width=1cm]{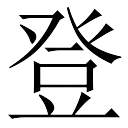}
        &\includegraphics[width=1cm]{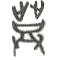}
        \includegraphics[width=1cm]{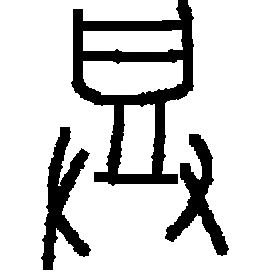}
        \includegraphics[width=1cm]{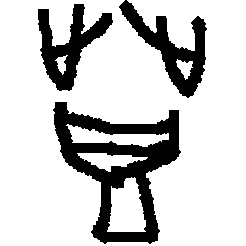}
        &\includegraphics[width=1cm]{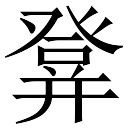}
        &\includegraphics[width=1cm]{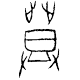}
        \includegraphics[width=1cm]{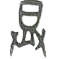}
        \includegraphics[width=1cm]{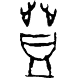}\\\hline
    \end{tabular}
    \caption{Comparison of oracle bone character annotations using Modern and Variant characters as category labels.}
    \label{tab:Merging}
\end{table}

\subsubsection*{Websites \& Databases}

The images of the OBCs collected from websites and databases have already been preprocessed by scanning, cropping, and alignment. Therefore, there is no need to design automatic annotation algorithms for this data, unlike the approach required for data from book sources. However, the following essential processes are still required:

\begin{enumerate}

    \item\textbf{Filtering:} It is important to note that a portion of the data on the GuoXueDaShi website, contributed by enthusiasts of ancient Chinese culture, cannot be fully guaranteed for reliability. 

    Specifically, these OBC images are of higher resolution and quality compared to other sources, but their unreliability stems from their labels. Currently, only about 1,500 categories of OBC have been deciphered, whereas the GuoXueDaShi website has 2,756 categories. This indicates that some undeciphered OBCs have been labeled by enthusiasts without expert verification, making them unreliable. 
    Consequently, in our filtration process, we cross-referenced these with other sources. This allowed us to identify 1,390 categories of OBC images that were unique to GuoXueDaShi and could not be verified. As a result, we retained only 1,366 out of the initial 2,756 categories after excluding these unverifiable samples. The samples of these 1390 categories, due to their lack of reliability, have not been classified as deciphered or undeciphered samples and are stored separately in the dataset.

    \item\textbf{Code Matching:} The OBC images from online and database sources are marked with specific codes, which we further mapped into modern Chinese characters. For the oracle bone inscriptions of YinQiWenYuan and HWOBC, the \dataset{} dataset only includes individual oracle bone characters, not compound characters. The term 'compound characters' refers to oracle bone characters corresponding to two or more words. Moreover, HWOBC is classified based on the character forms, leading to multiple character forms corresponding to the same Chinese character. Here, we merge them into the same category based on the corresponding Chinese character.
\end{enumerate}

\begin{figure}[t!]
    \centering
    \includegraphics[width=0.5\textwidth]{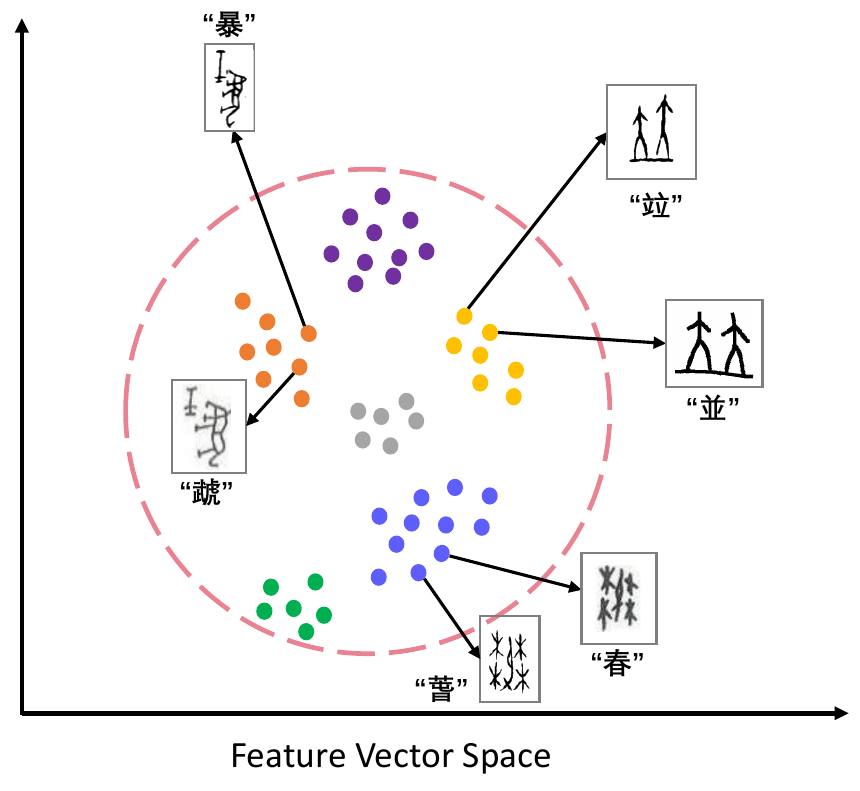}
    \caption{Oracle bone character images classified into the redundant category exhibit higher similarity in the feature vector space.}
    \label{fig:similarity}
\end{figure}

\subsection*{Integration}
 
In the stages of \textit{Data Acquisition and Automatic Annotation}, images of OBCs from distinct sources were collected and annotated. However, it is important to note that the annotation conventions for one OBC may vary depending on the source. For instance, as shown in Table~\ref{tab:Merging}, some sources might use standard modern Chinese characters for annotation, while other sources may prefer using corresponding Variant Chinese characters~\cite{bokset2006long} (\href{https://en.wikipedia.org/wiki/Variant\_Chinese\_characters}{https://en.wikipedia.org/wiki/Variant\_Chinese\_characters})
 for annotation. This leads to a scenario where images of OBCs that should belong to the same category are classified into different categories, creating redundant categories. Table~\ref{tab:Merging} illustrates this with examples of duplicate annotations, where each row shows how the same OBC image is categorized differently under the \textit{Modern Character Category} and the \textit{Variant Character Category}. To eliminate these redundancies, we integrate the data from different sources. For this purpose, we trained a widely-used unsupervised visual representation learning model MoCo~\cite{he2020momentum}, with OBC images from all sources. Subsequently, all the oracle bone images were encoded into a feature vector by the model. As illustrated in Figure~\ref{fig:similarity}, by calculating the similarity of these feature vectors, we merged similar samples into the same categories. In this way, we were able to reduce the original 1,781 categories obtained from different sources to 1,588, eliminating redundant categories.

\subsection*{Validation}

After undergoing all the procedures, we obtained a preliminary dataset. However, due to potential errors that might occur in the automated data acquisition and annotation process, we enlisted the expertise of OBC scholars from Anyang Normal University to meticulously review our dataset. Using authoritative books and the HWOBC database fonts as reference standards, they compared and evaluated OBC data in the HUST-OBC, discarding samples with errors and retaining the relatively accurate ones. This review produced the \dataset{} dataset.

\section*{Data Records}

\begin{table}[t]
\centering
\small
\begin{tabular}{lc|cc|cc}
\hline
\multicolumn{1}{l|}{\multirow{2}{*}{Source}} & \multirow{2}{*}{Type} & \multicolumn{2}{c|}{\#Categories}              & \multicolumn{2}{c}{\#Samples} \\ \cline{3-6} 
\multicolumn{1}{l|}{} & & \multicolumn{1}{c|}{Deciphered} & Undeciphered & \multicolumn{1}{c|}{Deciphered} & Undeciphered \\ \hline
\multicolumn{1}{l|}{New Compilation of Oracle Bone Scripts} & Book & \multicolumn{1}{c|}{1,203} & 2,288 & \multicolumn{1}{c|}{17,609} & 6,494 \\
\multicolumn{1}{l|}{Oracle Bone Script: Six Digit Numerical Code} & Book & \multicolumn{1}{c|}{1,148}      & 4,444 & \multicolumn{1}{c|}{9,609} & 4,444        \\
\multicolumn{1}{l|}{YinQiWenYuan} & Website & \multicolumn{1}{c|}{1,164} & 2,363 & \multicolumn{1}{c|}{1,697} & 2,363 \\
\multicolumn{1}{l|}{GuoXueDaShi} & Website & \multicolumn{1}{c|}{1,366} & 0 & \multicolumn{1}{c|}{16,259} & 0 \\
\multicolumn{1}{l|}{HWOBC} & Database & \multicolumn{1}{c|}{1,122} & 2,315 & \multicolumn{1}{c|}{31,890} & 49,688 \\ \hline
\multicolumn{2}{l|}{Total} & \multicolumn{1}{c|}{1,588} & 9,411 & \multicolumn{1}{c|}{77,064} & 62,989 \\ \hline
\end{tabular}
\caption{Statistic of oracle bone character data collected from various sources.}
\label{tab:data-source}
\end{table}

\begin{figure}[t!]
    \centering    \includegraphics[width=0.8\textwidth]{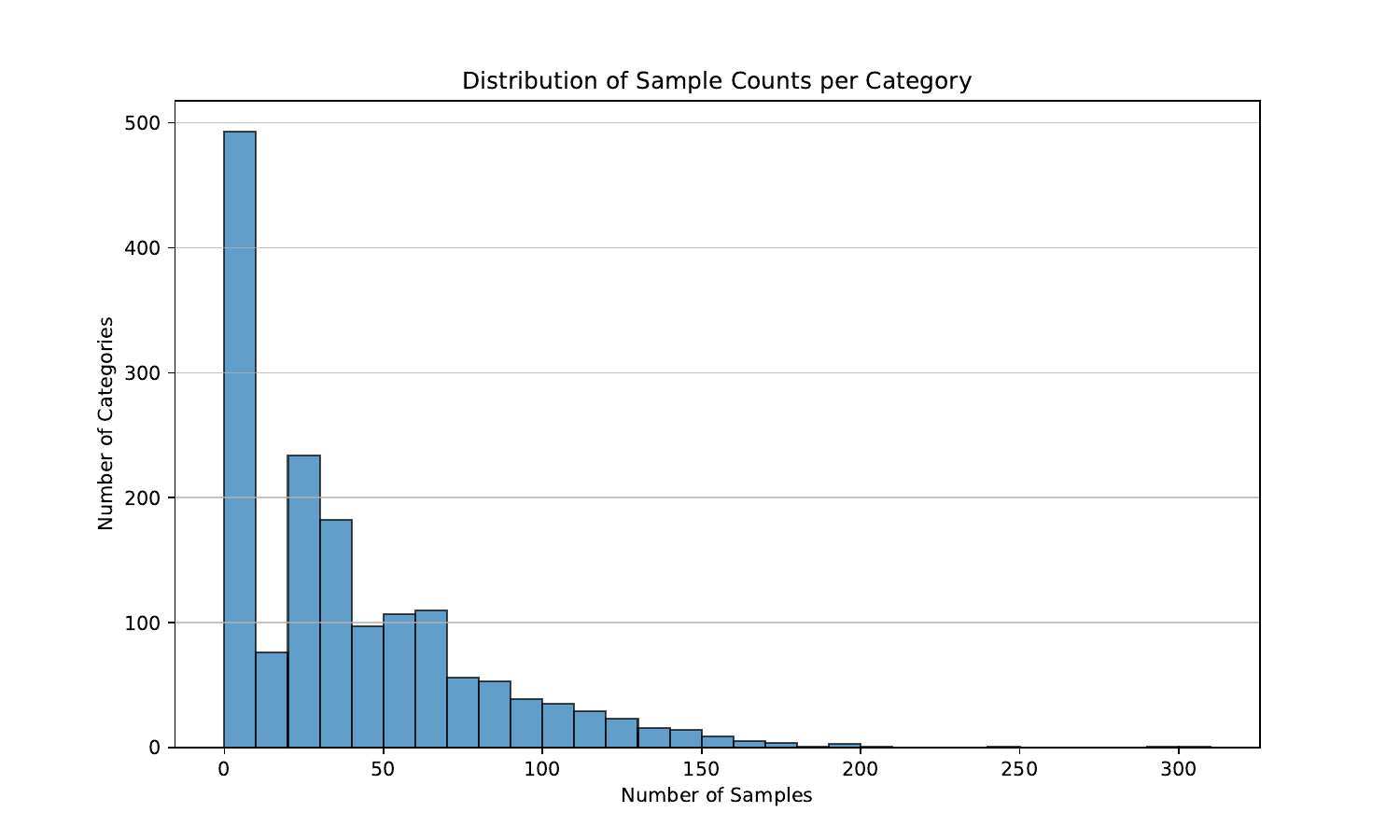}
    \caption{Histogram illustrating the distribution of sample counts across various categories in the \dataset{} deciphered split.}
    \label{fig:distribution}
\end{figure}

The \dataset{}\cite{HUSTOBC} comprises a total of 140,053 images sourced from five different origins, divided into deciphered and undeciphered sections. The deciphered section contains 77,064 images spanning 1,588 categories of individual characters, and the undeciphered section features 62,989 images across 9,411 categories of characters. Due to the lack of annotations for undeciphered categories, there may be duplicates among these 9,411 categories of undeciphered OBCs, which can only be merged once they are deciphered. Table~\ref{tab:data-source} provides detailed statistics of the OBC images obtained from these sources. Additionally, Figure~\ref{fig:distribution} presents a distribution histogram showing the number of sample images per category in our dataset. It reveals that most categories have fewer than 10 sample images, with the largest category boasting over 300 images.

\begin{figure}[t!]
    \centering
    \includegraphics[width=0.85\linewidth]{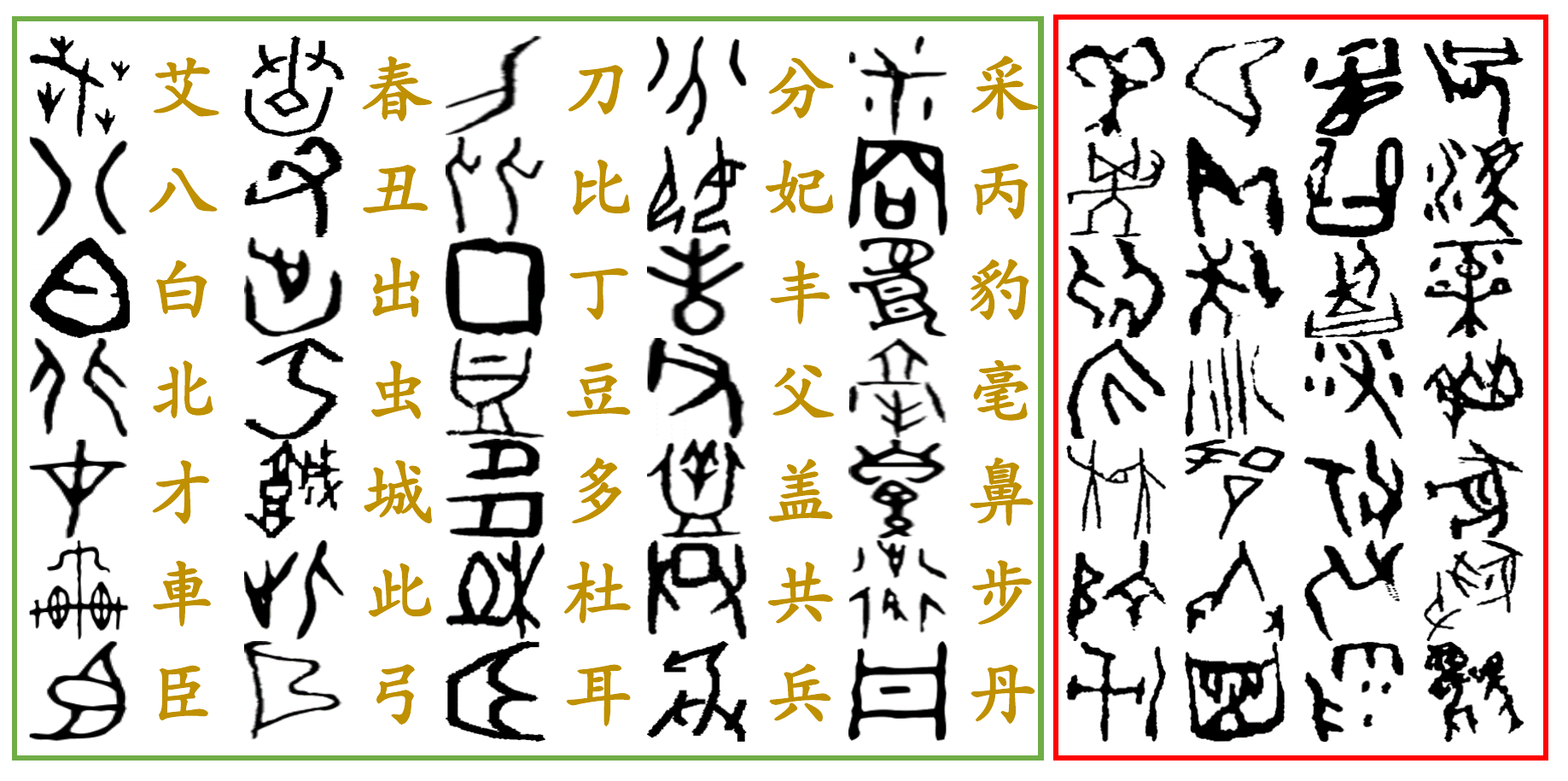}
    \caption{Samples of \textcolor{green}{\textbf{deciphered}} and \textcolor{red}{\textbf{undeciphered}} Oracle Bone Characters from the \dataset{} dataset.}
    \label{fig:deciphered-undeciphered-samples}
\end{figure}

For efficient retrieval, the \dataset{} is organized and stored by category names. Each image file is systematically named following the format <source>\_<label>\_<filename>, encapsulating its origin, category number, and sequence number, and is stored in folders named after their category number. For the deciphered categories, we have corresponding category numbers and the corresponding Chinese dictionary stored in a UTF-8 encoded JSON file. Figure~\ref{fig:deciphered-undeciphered-samples} demonstrates some deciphered and undeciphered OBCs from the \dataset{}.

\section*{Technical Validation}

One of the primary objectives in creating the \dataset{} is to facilitate future AI-assisted tasks in deciphering OBCs. To this end, we further assessed the quality of the dataset by employing it to train AI models. Specifically, we divided the deciphered section of the \dataset{} dataset into a training set, a validation set, and a test set using stratified sampling with proportions of 8:1:1, using them for training, validation, and testing in image classification tasks. Due to the limitation of classification models not being able to categorize unseen classes, we allocated all classes with only one sample into the training set. The accuracy of image classification can reflect the quality of the dataset to some extent. If the images in the dataset are of poor quality or have many labeling errors, the classifier's accuracy will be low, and vice versa. We employed the widely-used ResNet-50~\cite{resnet} as the backbone network for training. We tested the test set using the model that achieved the highest accuracy on the validation set, ultimately achieving a classification accuracy of 94.6\% and a macro-average F1 score of 0.914, which validates the dataset's quality and potential academic value. Table~\ref{tab:Validation} shows the model's recognition accuracy in some categories and provides example input images from different sources.

\section*{Licenses}
The dataset is released under a non-commercial license, CC BY-NC 4.0 (\href{https://creativecommons.org/licenses/by-nc/4.0/deed.en}{https://creativecommons.org/licenses/by-nc/4.0/deed.en}), which permits users to reuse and reproduce the dataset for research purposes.
\section*{Usage Notes}

The \dataset{} is available as a compressed archive, comprising three distinct folders. These folders separately house images of OBCs. The first one is for those that have already been deciphered, the second one is for those still awaiting interpretation, and the third one is for unreliable data from GuoXueDaShi. Within each folder, subfolders are organized by categories, containing images of OBCs corresponding to their respective categories. For more information, please see (\href {https://github.com/Pengjie-W/HUST-OBC}{https://github.com/Pengjie-W/HUST-OBC}).
\section*{Code Availability}

\noindent OpenCV toolkit is used to detect the borders in scanned book pages, which is available at \href{https://opencv.org/}{https://opencv.org/}.

\noindent The models and code for Chinese OCR, MoCo, and ResNet50 for Validation are available at (\href {https://github.com/Pengjie-W/HUST-OBC}{https://github.com/Pengjie-W/HUST-OBC}).

\begin{table}[t!]
    \centering
    \begin{tabular}{|c|c|c|c|c|c|c|c|}
    \hline
    &&\multicolumn{5}{c|}{Sample Input from Different Sources} & \\\cline{3-7}
    Ground-truth Category & \#Samples & A & B & C & D & E & Test Accuracy \\\hline
    \includegraphics[width=1cm]{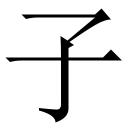}&307
    &\includegraphics[width=1cm]{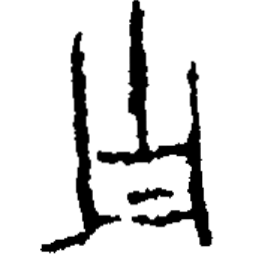}
    &\includegraphics[width=1cm]{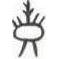}
    &\includegraphics[width=1cm]{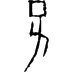}
    &\includegraphics[width=1cm]{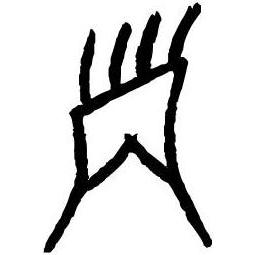}
    &\includegraphics[width=1cm]{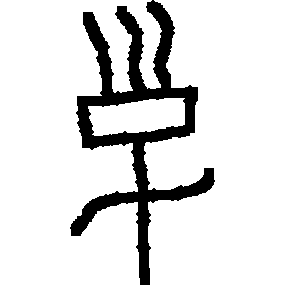}&90.32\%\\\hline
    \includegraphics[width=1cm]{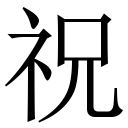}&295
    &\includegraphics[width=1cm]{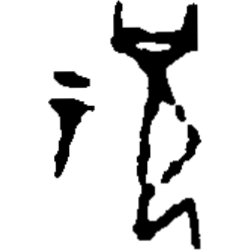}
    &\includegraphics[width=1cm]{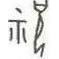}
    &\includegraphics[width=1cm]{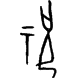}
    &\includegraphics[width=1cm]{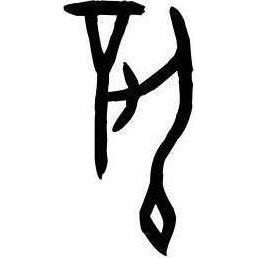}
    &\includegraphics[width=1cm]{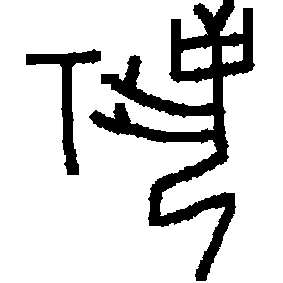}&93.10\%\\\hline
    \includegraphics[width=1cm]{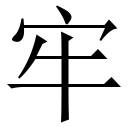}&209
    &\includegraphics[width=1cm]{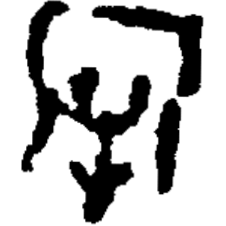}
    &\includegraphics[width=1cm]{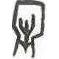}
    &\includegraphics[width=1cm]{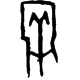}
    &\includegraphics[width=1cm]{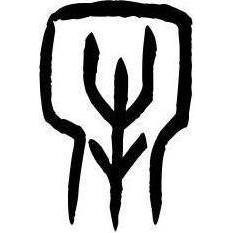}
    &\includegraphics[width=1cm]{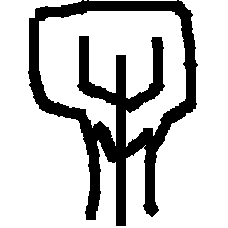}&100.00\%\\\hline
    \includegraphics[width=1cm]{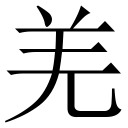}&199
    &\includegraphics[width=1cm]{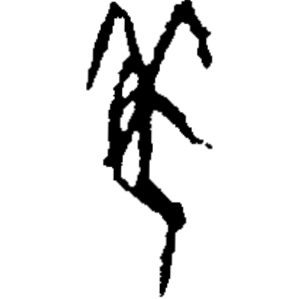}
    &\includegraphics[width=1cm]{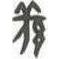}
    &\includegraphics[width=1cm]{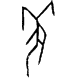}
    &\includegraphics[width=1cm]{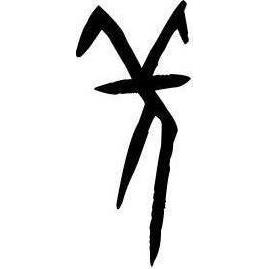}
    &\includegraphics[width=1cm]{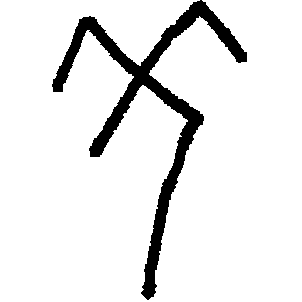}&100.00\%\\\hline
    \includegraphics[width=1cm]{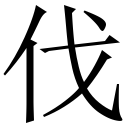}&190
    &\includegraphics[width=1cm]{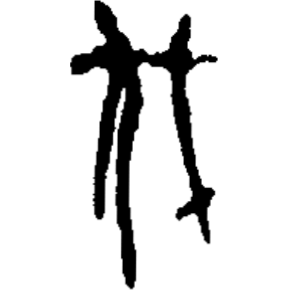}
    &\includegraphics[width=1cm]{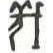}
    &\includegraphics[width=1cm]{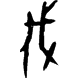}
    &\includegraphics[width=1cm]{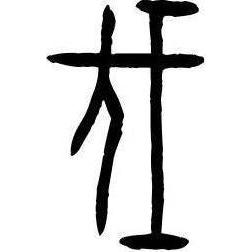}
    &\includegraphics[width=1cm]{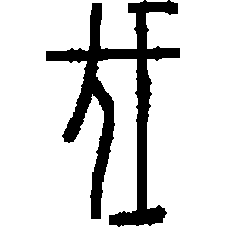}&100.00\%\\\hline
    \includegraphics[width=1cm]{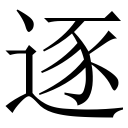}&180
    &\includegraphics[width=1cm]{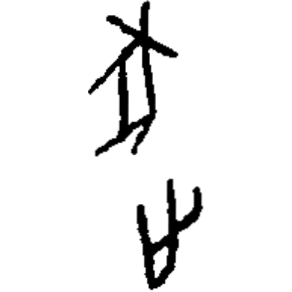}
    &\includegraphics[width=1cm]{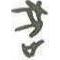}
    &\includegraphics[width=1cm]{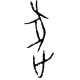}
    &\includegraphics[width=1cm]{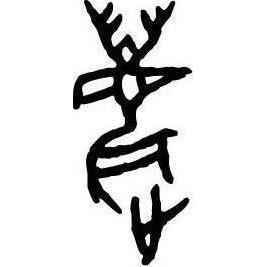}
    &\includegraphics[width=1cm]{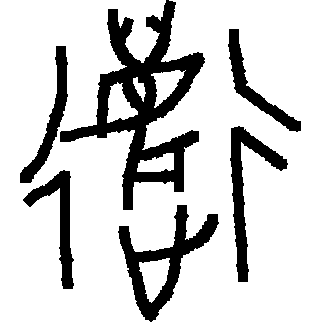}&94.44\%\\\hline
    \includegraphics[width=1cm]{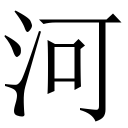}&179
    &\includegraphics[width=1cm]{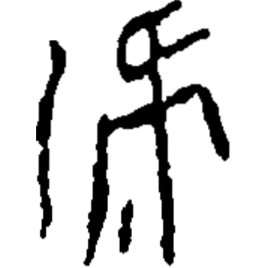}
    &\includegraphics[width=1cm]{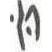}
    &\includegraphics[width=1cm]{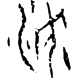}
    &\includegraphics[width=1cm]{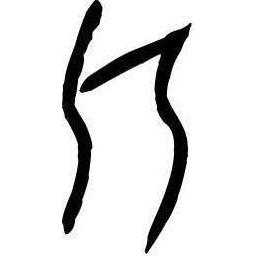}
    &\includegraphics[width=1cm]{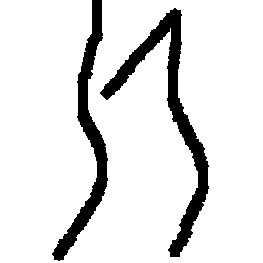}&100.00\% \\\hline
    \includegraphics[width=1cm]{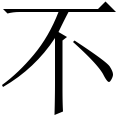}&106
    &\includegraphics[width=1cm]{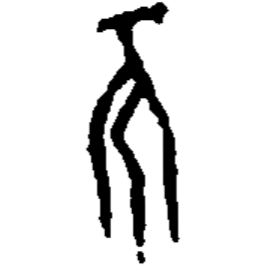}
    &\includegraphics[width=1cm]{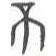}
    &\includegraphics[width=1cm]{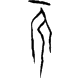}
    &\includegraphics[width=1cm]{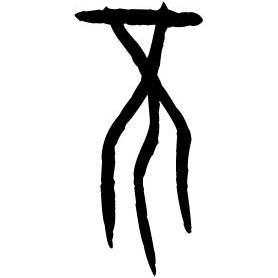}
    &\includegraphics[width=1cm]{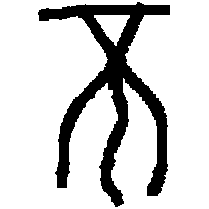}&80.00\%\\\hline
    \includegraphics[width=1cm]{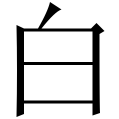}&93
    &\includegraphics[width=1cm]{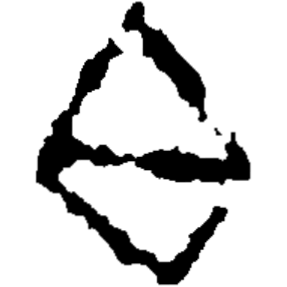}
    &\includegraphics[width=1cm]{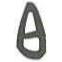}
    &\includegraphics[width=1cm]{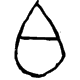}
    &\includegraphics[width=1cm]{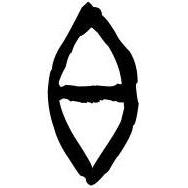}
    &\includegraphics[width=1cm]{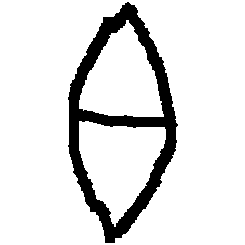}&100.00\%\\\hline
    \includegraphics[width=1cm]{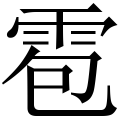}&63
    &\includegraphics[width=1cm]{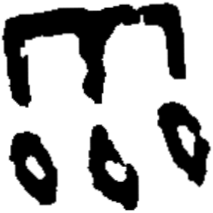}
    &\includegraphics[width=1cm]{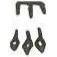}
    &\includegraphics[width=1cm]{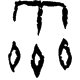}
    &\includegraphics[width=1cm]{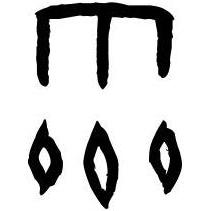}
    &\includegraphics[width=1cm]{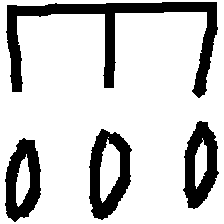}&100.00\%\\\hline
    \includegraphics[width=1cm]{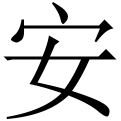}&55
    &\includegraphics[width=1cm]{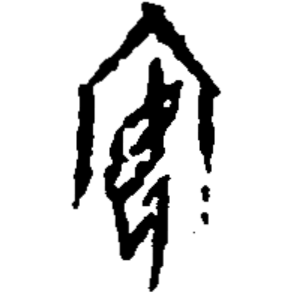}
    &\includegraphics[width=1cm]{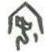}
    &\includegraphics[width=1cm]{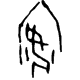}
    &\includegraphics[width=1cm]{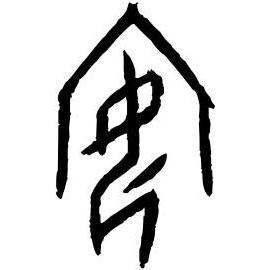}
    &\includegraphics[width=1cm]{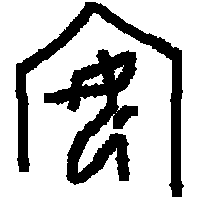}&100.00\%\\\hline
    \end{tabular}
    \caption{Validation accuracy of ResNet-50 on selected categories of the \dataset{} dataset.}
    \label{tab:Validation}
\end{table}

% \bibliography{ref}

\section*{Acknowledgements} %(not compulsory)

The authors thank the Key Laboratory of Oracle Bone Script Information Processing, Ministry of Education, Anyang Normal University for providing ancient text data sources and review of the dataset construction. 

\section*{Author contributions statement}

Pengjie Wang conducted experiments on Automatic Annotation, Integration, and Technical Validation. Haisu Guan, Jinpeng Wan, Pengjie Wang, and Kaile Zhang collectively obtained data from books. Pengjie Wang and Kaile Zhang collectively crawled data from websites. Xinyu Wang analyzed the results. Shengwei Han and Yongge Liu, as oracle bone script experts, supervised and assisted in the establishment of the dataset. Yuliang Liu provided guidance on the entire project. Xiang Bai provided laboratory resources. All authors reviewed the manuscript. 

\section*{Competing interests} %(mandatory statement)

The corresponding author is responsible for providing a \href{https://www.nature.com/sdata/policies/editorial-and-publishing-policies#competing}{competing interests statement} on behalf of all authors of the paper. This statement must be included in the submitted article file.

\end{document}